\documentclass{article}
\usepackage{natbib}
\usepackage{amsmath,graphicx}
\usepackage{booktabs}
\usepackage{soul, color}
\usepackage{float}
\usepackage[utf8]{inputenc}
\usepackage{authblk}
\usepackage{placeins}
\usepackage[top=1in, bottom=1in, left=1in, right=1in]{geometry} 
\usepackage{makeidx}  
\usepackage[utf8]{inputenc}
\usepackage{graphicx}
\usepackage[misc]{ifsym}
\usepackage{comment}
\usepackage{multirow}
\usepackage{color,soul}
\usepackage{amssymb}

\usepackage{bm}
\DeclareMathOperator*{\argmax}{argmax} 
\usepackage{tabularx}

\usepackage{hyperref}
\hypersetup{
    colorlinks=true,
    linkcolor=black,
    filecolor=black,      
    urlcolor=black,
    citecolor=black
}

\title{Collaborative Attention Mechanism for Multi-View Action Recognition}
\author[1]{Yue Bai \thanks{bai.yue@northeastern.edu}}
\author[2]{Zhiqiang Tao\thanks{ztao@scu.edu}}
\author[1]{Lichen Wang\thanks{wanglichenxj@gmail.com}}
\author[3]{Sheng Li\thanks{sheng.li@uga.edu}}
\author[1]{Yu Yin\thanks{yin.yu1@northeastern.edu}}
\author[1]{Yun Fu\thanks{yunfu@ece.neu.edu}}
\affil[1]{Department of Electrical and Computer Engineering, Northeastern University, USA}
\affil[2]{Department of Computer Science and Engineering,
Santa Clara University, USA}
\affil[3]{Department of Computer Science, University of Georgia, USA}

\date{}

\begin{document}
\maketitle

\begin{abstract}
Multi-view action recognition (MVAR) leverages complementary temporal information from different views to improve the learning performance. Obtaining informative view-specific representation plays an essential role in MVAR. Attention has been widely adopted as an effective strategy for discovering discriminative cues underlying temporal data. However, most existing MVAR methods only utilize attention to extract representation for each view individually, ignoring the potential to dig latent patterns based on mutual-support information in attention space. To this end, we propose a collaborative attention mechanism (CAM) for solving the MVAR problem in this paper. The proposed CAM detects the attention differences among multi-view, and adaptively integrates frame-level information to benefit each other. Specifically, we extend the long short-term memory (LSTM) to a Mutual-Aid RNN (MAR) to achieve the multi-view collaboration process. CAM takes advantages of view-specific attention pattern to guide another view and discover potential information which is hard to be explored by itself. It paves a novel way to leverage attention information and enhances the multi-view representation learning. Extensive experiments on four action datasets illustrate the proposed CAM achieves better results for each view and also boosts multi-view performance.
\end{abstract}

\section{Introduction}
Multi-view action recognition (MVAR) has drawn more attention~\cite{cai2014multi, cheng2012human, holte20113d} since the increasing usage of multi-modality sensors to improve recognition performance. It is a challenging task due to the difficulties of extracting complicated temporal patterns and fully utilizing multi-view information. Existing MVAR scenarios can be roughly grouped into two categories. The first category captures multi-view actions using multiple sensors belonging to the same modality (e.g., RGB cameras). These sensors are deployed in different viewpoints (e.g., front, top, and side), and the corresponding methods~\cite{cai2014multi, holte20113d, ji2014view} are designed to handle these geometric views. The second category collects multi-view actions via different modality sensors (e.g., RGB, depth, and skeleton)~\cite{lin2012human, rahmani2016histogram, wang2019ev}. For example, Kinect sensors collect RGB, depth, and skeleton data simultaneously~\cite{pagliari2015calibration, zhang2012microsoft}. Some modalities containing physical characteristics, e.g., electromyography (EMG) and acceleration, are also used to explore MVAR~\cite{bu2009hybrid, chen2015utd, de2009guide}. Generally, various modalities could provide complementary information to achieve better performance than a single modality.

\begin{figure}[t]
\centering
\begin{center}
\includegraphics[width=0.7\linewidth]{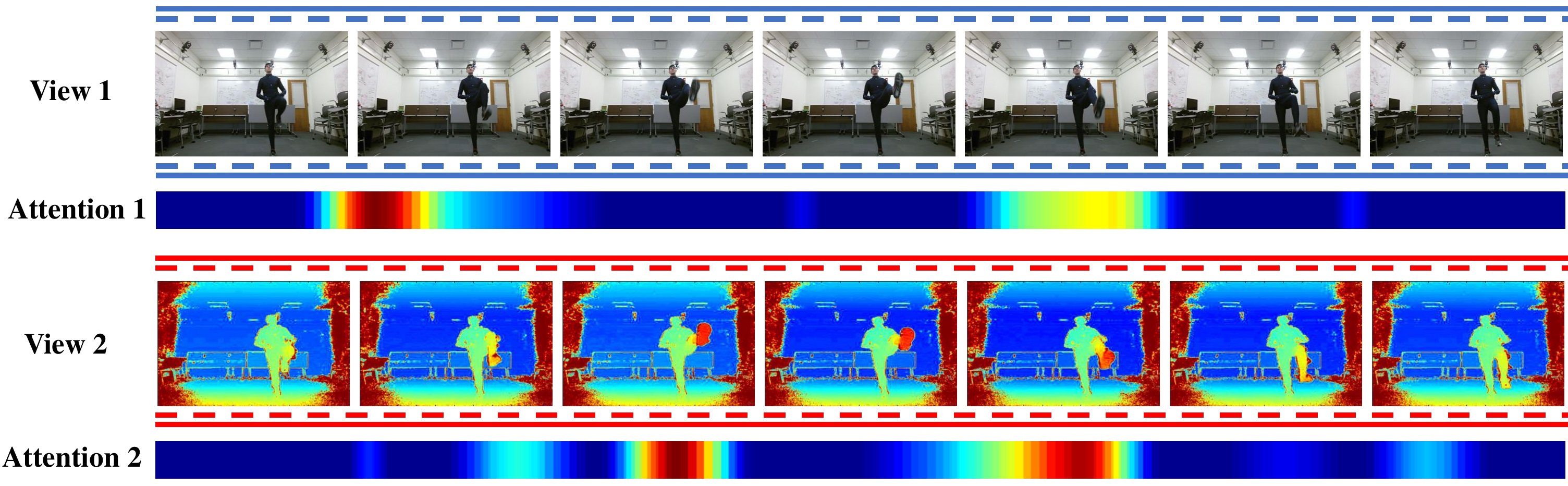}
\end{center}
\vspace{-2mm}
\caption{Illustration of the difference among multiple views in attention space, where the RGB view easily captures visible changes between images, and the depth view is more sensitive to in-depth distance changes. We take ``kicking'' action as an example. In RGB view, the visible changes are obvious during the lifting leg and drawing back the leg. These frames are easily captured by RGB view as patterns. However, in the middle of example sequence, the changes of RGB view are tiny and hard to be discovered, whereas the depth view illustrates significant changes during this period (when the leg is at the top position, the changes are mainly in depth dimension).}
\label{fig:motivation}\vspace{-2mm}
\end{figure}

In this paper, we focus on the second category for the MVAR task. Its main challenges lie in \emph{1) how to efficiently represent view-specific information} and \emph{2) how to utilize them for achieving better multi-view performance}. Subspace learning is widely used to seek a common subspace for multiple views~\cite{jia2016low, jia2014latent}. It aims to find consistent characteristics among multi-view and derive effective representations for recognition. However, emphasizing the synchronous patterns may overlook the distinctive information of each view. Fusion mechanism is another popular way for multi-view learning~\cite{wang2019generative, zadeh2018memory, wang2019icdm}. While effective fusion could take advantage of each view's distinctive information and combine them for encouraging higher performance, some straightforward fusion methods (e.g., average, concatenation, and summation) may hurt the final performance. On the one hand, late fusion algorithms fully explore distinctive features from each view individually and focus on fusion in label space~\cite{wang2019icdm, bruno2009multiview}. The mutual-support information across views is utilized by wisely fusing the predicted scores. On the other hand, early fusion methods pay more attention on augmenting the capacity of each view by borrowing information from the other views~\cite{zadeh2018memory, liang2018multimodal}. They integrate the multi-view information in feature space. However, both late and early fusion strategies benefit multi-view learning only by exploiting the readily available mutual-support information directly. For example, the late fusion uses the predicted scores obtained from view-specific classifier; the early fusion borrows information from each other. 

\begin{figure*}[t]
\centering
\begin{center}
\includegraphics[width=1.0\linewidth]{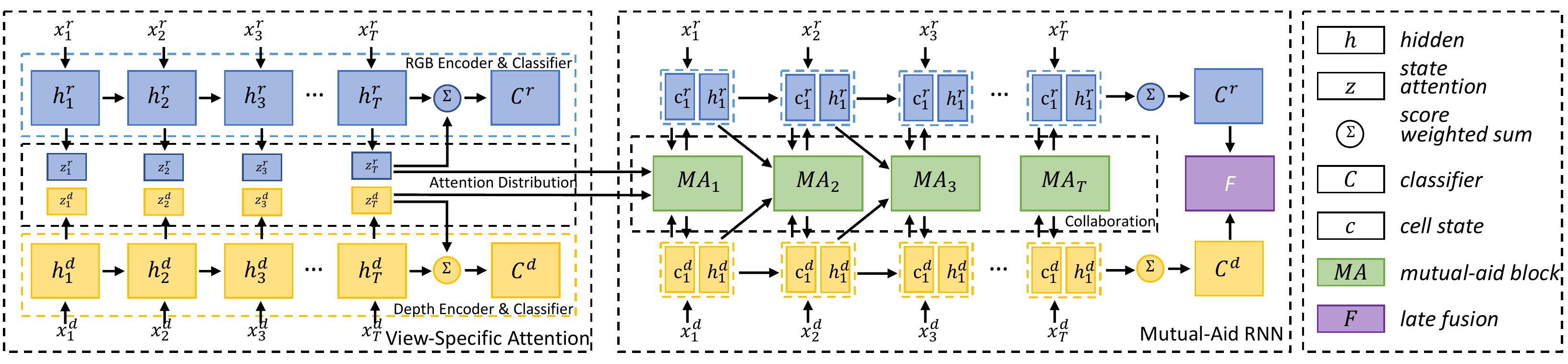}
\end{center}
\vspace{-2mm}
\caption{Illustration of our whole framework. The view-specific attention is the first-stage. Two LSTM encoders make view-specific recognition, respectively. The attention score $z_t$ of two views are obtained and fed into the second stage. Mutual-Aid RNN, as the second stage, achieves the collaborative attention mechanism. Mutual-Aid block, \textit{MA}$_t$ collaborate the two views frame-by-frame. We deploy attention module for each view and make a fusion for multi-view results.}
\label{fig:framework}
\end{figure*}

To further discover the latent cues across multiple views, we propose a collaborative attention mechanism (CAM) model for multi-view sequential learning as shown in Fig.~\ref{fig:framework}. Attention is an effective mechanism to enhance the representation learning with the capacity of interpreting model and providing intuitions of data. Inspired by the interpretability of temporal attentions, we instantiate CAM based on the observations from multi-view actions: different views have different attention distributions (see Fig.~\ref{fig:motivation}). Specifically, the RGB view pays attention to certain frames, while the depth view values more contributions from some other frames. Each view has its own concentrations, yet ignoring the frames that are hard to explored by itself. To disclose the overlooked information, we propose a Mutual-Aid RNN (MAR) cell to collaboratively guide multi-view representation learning. Based on the attention differences, one view utilizes the attention differences and selectively directs the other view to focus on certain frames containing obscure information. We summarize the contribution of this work in the following.
\begin{itemize}
\item We propose a collaborative attention mechanism (CAM) framework to improve the MVAR performance. It efficiently utilizes the attention information across different views to mutually enhance multi-view learning, which boosts the single-view and multi-view performance simultaneously. 
\item A novel Mutual-Aid RNN (MAR) cell is proposed for multi-view sequential data. It relies on attention distribution to capture the latent patterns and adaptively enhances the frame-level representations of each view.

\item We provide a new perspective to reacquaint multi-view learning by leveraging the interpretability of attention mechanism to guide learning process. Extensive experiments on four action datasets illustrate the effectiveness of the proposed CAM.
\end{itemize}
\section{Related Work}
Numerous multi-view learning algorithms have been proposed for multi-view action recognition (MVAR)~\cite{junejo2008cross, wang2014cross, wang2019generative}. Attention mechanism is popular in modeling sequential data, especially for human actions. In this paper, we aim to propose a novel Collaborative Attention Mechanism (CAM) for MVAR. To this end, we mainly focus on introducing multi-view learning algorithms and a variety of attention based methods.

\subsection{Multi-View Learning}
Multi-view learning technique has been adopted in multiple applications~\cite{jones2003fast, farfade2015multi, zhang2014cooperative, nie2017auto, bai2019multi, liu2015low} (e.g., image classification, emotion recognition, and face detection). Action recognition task is also expressively developed by multi-view learning~\cite{cai2014multi, ahmad2006hmm, shao2016kernelized}.
A generative model is proposed for MVAR which expands model to handle missing view scenario~\cite{wang2019generative}.
A generative feature fusion strategy is designed for human action recognition~\cite{wang2018pm}.
A multi-view super vector is proposed to fuse action feature descriptors for MVAR~\cite{cai2014multi}. However, these method only focus on using complementary information directly, while our CAM aims at discovering additional latent knowledge to benefit each single view and further boost multi-view performance.

\subsection{Attention Mechanism}
The pioneering work~\cite{mnih2014recurrent} introduces the attention mechanism into computer vision community to make image classification using recurrent model. The natural language processing (NLP) field was significantly explored by firstly employing attention on machine language translation task~\cite{bahdanau2014neural}. After that, attention based methods were widely adopted into many other applications (e.g., image caption~\cite{xu2015show}, text classification~\cite{yang2016hierarchical}, and image generation~\cite{gregor2015draw}). Benefit from the inherent advantage of attention for modeling sequential data, plenty works are proposed for action recognition. 
A visual attention framework is proposed with a soft attention based model for action recognition~\cite{sharma2015action}.
A spatial-temporal attention model is designed based on skeleton data for recognizing human action~\cite{song2017end}.
However, these works only use attention to boost performance instead of taking advantage of its instructive function to explore latent information and achieve further improvement.

\section{Methodology}


\subsection{Preliminary}
Let $X^1\in\mathbb{R}^{T \times d^1}$ and $X^2\in\mathbb{R}^{T \times d^2}$ are multi-view feature inputs. $T$ represents the length of action clip. $d^1$ and $d^2$ are feature dimensions of two views. $Y \in \mathbb{R}^{C}$ is the one-hot label matrix, where $C$ is the number of classes.

Before introducing each model component, we briefly summarize the whole logic of our proposed CAM. The first phase contains view-specific encoders and classifiers. We use LSTM with self-attention to encode the sequence input and obtain the attention information. The training is supervised by the label information. In the second phase, our CAM utilizes the view-specific attention distributions from the first phase to guide the multi-view collaboration learning. 
The view-specific representation is enhanced to achieve higher single view performance. The correlative late fusion is deployed to obtain multi-view result which is boosted by the enhanced single-view representations.

\subsection{Temporal Attention for Action Recognition}
Given an action sample and the corresponding label, the temporal attention model aims to encode the sequential input and optimize the following objective:

\begin{equation}\label{eq:temporal_attention_obj}
\theta^* = \argmax_{\theta} \sum_{(X, y)} \log p(y|X;\theta),
\end{equation}
where $\theta$ is the set of parameters of model. $X = \{ x_1, ..., x_t\}$ is the multiple frames of a video sample, and $y$ is the corresponding label. The dynamic information is the key factor for recognition. Thus, wisely choosing temporal encoder is decisive for temporal feature extraction. In our work, we deploy 
long short-term memory (LSTM)~\cite{hochreiter1997long} to model sequential data. Each input frame $x_t$ is encoded as a hidden representation $h_t$, and the cell state $c_t$ is updated correspondingly. The update processes in LSTM are given by
\begin{equation}\label{eq:LSTM_block}
\begin{aligned}
    f_t & = \sigma_{g}(W_f x_t + U_f h_{t-1} + b_f), \\
    i_t & = \sigma_{g}(W_i x_t + U_i h_{t-1} + b_i), \\
    o_t & = \sigma_{g}(W_o x_t + U_o h_{t-1} + b_o), \\ 
    c_t & = f_t \circ c_{t-1} + i_t \circ \sigma_c (W_c h_t + U_c h_{t-1} + b_c), \\
    h_t & = o_t \circ \sigma_h(c_t),
 \end{aligned}
\end{equation}
where $f_t$, $i_t$, $o_t$, $c_t$, and $h_t$ represent forget gate, input gate, output gate, cell state and hidden state at time $t$. $c_{t-1}$ and $h_{t-1}$ are cell and hidden states at time $(t-1)$. $\sigma_g$, $\sigma_c$, and $\sigma_h$ are activation functions. $\circ$ represents the element-wise product. $W$, $U$ and $b$ are learnable parameters.

Original video sequence $X$ is encoded as $H = \{h_1, ..., h_T\}$. Commonly, we pick the last hidden state $h_T$ to represent the whole sequence.
However, it may lose temporal information to some degree. A reasonable way is using the weighted summation of $h_t$. The weights are calculated based on the importance of each frame by attention mechanism. Here, we adopt a self-attention variant~ \cite{yang2016hierarchical} which is proposed for document classification. It can be easily utilized for modeling temporal data and given by 
\begin{equation}\label{eq:self_attention}
\begin{aligned}
    u_t & = tanh(W_w h_t + b_w), \\
    z_t & = \frac{exp(u_t^T u_w)}{\sum_t exp(u_t^T u_w)},\\
    r & = \sum_t z_t h_t,
\end{aligned}
\end{equation}
where $u_t$ denotes the attention vector derived from $h_t$. $W_w$ and $b_w$ are learnable parameters. $u_w$ is the context vector, which is random initialized and updated through the optimization procedure. It depicts the global meaning of the video sequence itself. $z_t$ means the degree of importance for each $u_t$ among the whole video context $u_w$ by using softmax activation. $r$ is the weighted summation of $h_t$. 

To introduce our CAM clearly, we go deeper here to provide more insights about LSTM. The key factor is the $c_t$. It reflects the memory states of the whole sequence. $f_t$ and $i_t$ update the $c_t$ internally through the \textit{forget} and \textit{input} procedures. The contents of \textit{forget} and \textit{input} are derived from current input $x_t$ and last hidden state $h_{t-1}$. The content of current hidden state $h_t$ is also extracted from $x_t$ and $h_{t-1}$, then filtered by $c_t$. All information flows cross several control gates center on the $c_t$. As the memory state, $c_t$ only records the temporal dynamic characteristic instead of specific domain knowledge. Own to this insight, we conclude that fully exploiting the cell state is decisive for temporal encoding. We will introduce our framework starting from the temporal attention and $c_t$. 


\subsection{View-Specific Attention Mechanism}
Multi-view actions contain mutual-support information for each other, while each view has its own distinctive patterns. To fully exploit the distinctive information from each view, we propose the view-specific attention mechanism formulated as follows:

\begin{equation}\label{eq:view_specific_encoder}
\begin{aligned}
    	H^v & = E^v(X^v, \phi^v_E),\\
    	r^v & = Q^v(H^v, \phi^v_Q),\\
    	\hat{Y}_a^v & = C^v(r^v, \phi^v_C),\\
\end{aligned}
\end{equation}
where superscript $v$ represents the view $v$. $E$ is LSTM module (Eq.~\ref{eq:LSTM_block}), encoding sequence $X$ into hidden sequence $H$. $Q$ is the attention module (Eq.~\ref{eq:self_attention}), transferring $H$ into weighted summation vector $r$. $C$ is the view-specific linear classifier, resulting in the predicted label $\hat{Y}_a$. $\phi^v_E$, $\phi^v_Q$, and $\phi^v_C$ are learnable parameters. They are optimized by minimizing following objective:
\begin{equation}\label{eq:view_specific_loss}
L^v_a = \ell(Y, \hat{Y}^v_a),
\end{equation}
where $\ell$ represents cross-entropy loss, $Y$ is the ground truth.

The goal of view-specific attention aims to derive the $Z^v = \{z^v_1, ..., z^v_T\}$ which is the intermediate product of Eq.~\ref{eq:self_attention} and preserves the view-specific dynamic patterns. 
We regard the view-specific attention as our first stage model. Multi-view attention distributions $Z^v$ are reserved for the second stage model. 

\subsection{Multi-View Collaboration by Mutual-Aid RNN}

\begin{figure}[t]
\centering
\begin{center}
\includegraphics[width=0.58\linewidth]{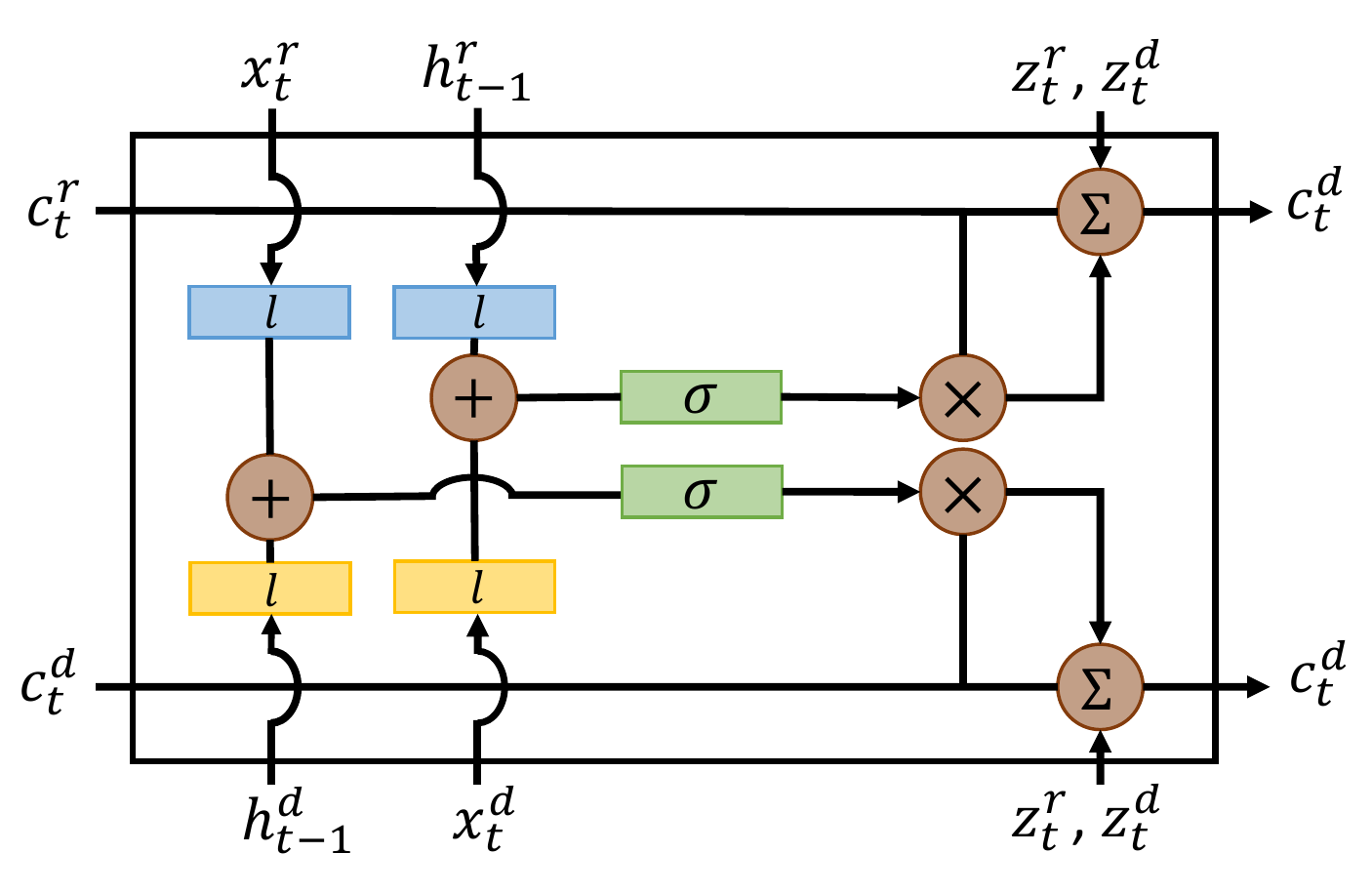}
\end{center}
\vspace{-5mm}
\caption{Illustration of the proposed Mutual-Aid RNN (MAR) cell. $x_t$ and $h_{t-1}$ of two views are set as input to integrate information from cross view. They update the $c_t$ of two views via point multiplication, respectively. Attention distributions $z_t$ of two views are involved in weighted summation to adaptively collaborate the multi-view knowledge. }
\label{fig:mar}
\vspace{-2mm}
\end{figure}

To substantially take advantage of multi-view data, in our second stage, we propose the multi-view collaboration mechanism. It mutually supports the multi-view learning. Note that unlike some data augmentation strategies cross multi-view (e.g., representation mapping, feature fusion, and generative model), our goal is borrowing the knowledge from other view to help the target view discovering more clues by itself, instead of transferring or fusing information directly. To convey our insight clearly, we briefly discuss it here. View-specific attention provides the distinctive temporal patterns by attention distributions. It is extracted through optimizing single view classifier individually and reflects the view-specific characteristics. Particularly, for the same action sample, the attention distribution of one view focuses on certain frames, while that of the other view focuses on different frames. This is caused by the inherent attribute of each view. For example, the RGB view always takes advantages of color changes to recognize human action, while the depth view focuses more on the distance changes. As a result, the effective frames for two views could be different. However, the differences are not opposite but complementary for each other. Some frames are ignored by certain view, due to its inherent attribute, still restore valuable information. This information may be easily discovered by the other view. To this end, we propose the multi-view collaboration mechanism. It encourages multi-view data to help with each other by guiding other view to focus on implicit but effective information. 

We first encode the multi-view action sequence $X^v$ with LSTM (Eq.~\ref{eq:LSTM_block}). We formulate it briefly as follows:
\begin{equation}\label{eq:second_LSTM_block}
\begin{aligned}
    c_t & = f_t \circ c_{t-1} + i_t \circ \sigma_c (W_c h_t + U_c h_{t-1} + b_c), \\
    h_t & = o_t \circ \sigma_h(c_t),
 \end{aligned}
\end{equation}
where $c_t$ and $h_t$ are the cell state and hidden state for time $t$, deriving information from $c_{t-1}$, $h_{t-1}$, and $x_t$. 

We extend the LSTM to a Mutual-Aid RNN (MAR) by designing a novel recurrent cell (see Fig.~\ref{fig:mar}). Instead of setting the $c_t$ as the cell state for next time step directly, MAR leverages the knowledge from the other view to update the $c_t$ of target view, and then guides the target view for information mining. Next, we formulate our proposed MAR step-by-step.

\textbf{Cross-View Collaborator} is proposed to integrate multi-view information and prepared for following collaborative learning. It is formulated as follows:

\begin{equation}\label{eq:mar_collaborator}
\begin{aligned}
    G_{r\rightarrow d} & = \sigma (W_{rd} x^r_t + W_d h^d_{t-1}), \\
    G_{d\rightarrow r} & = \sigma (W_{dr} x^d_t + W_r h^r_{t-1}), \\
 \end{aligned}
\end{equation}
where $W_{*}$ are learnable parameters. $G_{*}$ extract information from current input $x_t^{*}$ and collaborate with last hidden state $h_{t-1}^{*}$ from the other view. $\sigma$ represents the sigmoid activation. The frame-level knowledge from the other view is reserved in $G_{*}$.

\textbf{Mutual Filtering} is designed based on cross-view collaborator above. Cell state $c_t^{*}$ contains temporal dynamic patterns for each view. It could be updated internally in LSTM (Eq.~\ref{eq:second_LSTM_block}). However, $c_t^{*}$ will only contain the memory information from single view and cannot take advantage of temporal patterns of the other view. Mutual filtering helps model update the $c_t^{*}$ using cross-view collaborator to derive knowledge from the other view, which is given by
\begin{equation}\label{eq:mar_filter}
\begin{aligned}
    {c_t^r}' & = G_{d \rightarrow r} \circ c_t^r, \\
    {c_t^d}' & = G_{r \rightarrow d} \circ c_t^d, \\
 \end{aligned}
\end{equation}
where $\circ$ is the point-wise product. $c_t^{*\prime}$ are the enhanced cell states containing mutual-support temporal information from the other view.

\textbf{Mutual Collaboration} is finally achieved by combining the attention distributions and two proposed modules above. Attention distributions $z_t^{*}$ reflect the frame importance of each single-view. Further, it also decides the information importance during updating $c_t^{*}$ in multi-view collaborative learning. We first normalize the attention scores by
\begin{equation}\label{eq:normalize_att}
\begin{aligned}
    {z_t^r}' = \frac{z_t^r}{z_t^r + z_t^d},\\
    {z_t^d}' = \frac{z_t^d}{z_t^r + z_t^d}.\\
 \end{aligned}
\end{equation}
The original cell state $c_t^{*}$ are updated by single view information, while $c_t^{*\prime}$ are updated by the cross-view collaborator $G_{*}$. $z_t^{*\prime}$ represent the importance of dynamic knowledge from different views. We integrate the multi-view information for updating cell states via the weighted summation:
\begin{equation}\label{eq:weighted_sum}
\begin{aligned}
    {c_t^r}'' & = {z_t^r}' c_t^r + {z_t^d}' {c_t^r}',\\
    {c_t^d}'' & = {z_t^d}' c_t^d + {z_t^r}' {c_t^d}',\\
 \end{aligned}
\end{equation}
where $c_t^{*\prime\prime}$ are the final cell states containing the dynamic knowledge from multi-view data. Through being the inputs for next time step, they bring the knowledge from the other view to overcome the inherent drawback of each single view. In this way, some implicit information could be discovered by each single view via the guidance from mutual collaboration.

So far, we have introduced the proposed multi-view collaboration via our MAR encoder. Its input and output are multi-view action sample and sequential representation, respectively. In order to fully utilize the discovered information via our collaboration mechanism, we reuse the self-attention (Eq.~\ref{eq:self_attention}) to obtain the final representation and make the view-specific recognition again similar to Eq.~\ref{eq:view_specific_encoder}. We briefly formulate these steps by

\begin{equation}\label{eq:mar_encoder}
\begin{aligned}
    	H^v_M & = E^v_M(X^v, \phi^v_{E_M}),\\
    	r^v_M & = Q^v_M(H^v_M, \phi^v_{Q_M}),\\
    	\hat{Y}^v_M & = C^v_M(r^v_M, \phi^v_{M}),\\
\end{aligned}
\end{equation}
where all the terms with subscript $M$ represents the similar meanings with Eq.~\ref{eq:view_specific_encoder} under our multi-view collaboration mechanism. We obtain another attention distribution $Z^v_M$ and the predicted label $\hat{Y}^v_M$ for multi-view. The learnable parameters are optimized by minimizing following loss: 
\begin{equation}\label{eq:second_view_specific_loss}
L^v_M = \ell(Y, \hat{Y}^v_M).
\end{equation}
The view-specific attention (first stage) and the multi-view collaboration (second stage) constitute our whole framework Collaborative Attention Mechanism (CAM). It exploits the knowledge from multi-view attention distributions to guide the multi-view information discovering and enhance the learning process. More implicit but valuable patterns could be discovered for performance boosting. After obtaining the $\hat{Y}^v_M$ from multi-view, we use a correlative late fusion to evaluate final multi-view results.  

\subsection{Correlative Late Fusion}
Our CAM discovers more clues to enhance the single view representation. 
We deploy a correlative late fusion~\cite{wang2019icdm} for multi-view evaluation, which is given by

\begin{equation}\label{eq:late_fusion}
\begin{aligned}
    D & = \hat{Y}_M^r \cdot \hat{Y}_M^{d\top}, \\
\end{aligned}
\end{equation}
where $\hat{Y}_M^r \in \mathbb{R}^{d^l \times 1}$ and $\hat{Y}_M^{d\top} \in \mathbb{R}^{1 \times d^l}$ are the predicted label from multi-view. $D \in \mathbb{R}^{d^l \times d^l}$ is the correlative matrix constructed by the multiplication of multi-view predicted labels. 
$D$ is flatten into a $d^l \times d^l$ dimension vector as input of the final classifier $C^f: \mathbb{R}^{d^l \times d^l} \rightarrow \mathbb{R}^{d^l}$. $C^f$ is parameterized by $\phi_{C^f}$ and updated by minimizing following loss:

\begin{equation}\label{eq:late_loss}
\begin{aligned}
    L_f = \ell(Y, C^f(D, \phi_{C^f})),\\
\end{aligned}
\end{equation}
where $Y$ is the ground truth, $\ell$ is the cross-entropy loss. $L_f$ represents the final multi-view loss. Our model mainly consists of the view-specific attention and the multi-view collaboration, followed by a late fusion model for multi-view learning performance.

As a summary, view-specific attention aims to capture the differences between two views, especially focusing on the attention distribution. These differences are leveraged as guidance information for multi-view collaboration. A novel MAR cell is proposed for extracting cross-view knowledge and updating memory cell effectively. More implicit yet valuable information could be discovered. A concise late fusion is deployed to evaluate multi-view performance.

\begin{table}[t]
\begin{center}
\scalebox{1.0}{
\begin{tabular}{llccc}
\toprule
Datasets & Methods & RGB & Depth & Fusion  \\
\toprule
\multirow{6}{*}{EV-Action}
& TSN~\cite{wang2016temporal} & 0.6855 & 0.6723 & -\\
& RC Classifier~\cite{bianchi2018reservoir} & 0.5992 & 0.5790 & 0.6213\\
& MFN~\cite{zadeh2018memory} & 0.5743 & 0.4082 & 0.6423  \\
& MLSTM-FCN~\cite{karim2019multivariate} & 0.6804 & 0.6926 & 0.7014 \\
& GMVAR~\cite{wang2019generative} & 0.6792 & 0.6739 & 0.7088\\
& CAM (ours) & \textbf{0.7022} & \textbf{0.7123} & \textbf{0.7359}\\
\midrule
\multirow{6}{*}{NTU} 
& TSN~\cite{wang2016temporal} & 0.7517 & 0.7691 & -  \\
& RC Classifier~\cite{bianchi2018reservoir} & 0.7683 & 0.8014 & 0.8258  \\
& MFN~\cite{zadeh2018memory} & 0.7089 & 0.8062 & 0.8125  \\
& MLSTM-FCN~\cite{karim2019multivariate} & 0.7662 & 0.7941 & 0.8217  \\
& GMVAR~\cite{wang2019generative} & 0.7545 & 0.7702 & 0.8018  \\
& CAM (ours) & \textbf{0.7720} & \textbf{0.8134} & \textbf{0.8408} \\
\midrule
\multirow{5}{*}{DHA} 
& TSN~\cite{wang2016temporal} & 0.6785 & 0.8324 & - \\
& AMGL~\cite{nie2016parameter} & 0.6461 & 0.7284 & 0.7489 \\
& MLAN~\cite{nie2017multi} & 0.6791 & 0.7296 & 0.7613 \\
& GMVAR~\cite{wang2019generative} & 0.6972 & 0.8348 & \textbf{0.8872} \\
& CAM (ours) & \textbf{0.7407} & \textbf{0.8642} & 0.8724 \\
\midrule
\multirow{5}{*}{UWA3D} 
& TSN~\cite{wang2016temporal} & 0.4833 & 0.5936 & - \\
& AMGL~\cite{nie2016parameter} & 0.3067 & 0.3667 & 0.3933 \\
& MLAN~\cite{nie2017multi} & 0.2933 & 0.2867 & 0.3800 \\
& GMVAR~\cite{wang2019generative} & 0.4917 & 0.5846 & 0.6035  \\
& CAM (ours) & \textbf{0.5083} & \textbf{0.6073} & \textbf{0.6314}  \\
\bottomrule
\end{tabular}}
\end{center}
\caption{Classification accuracy on four multi-view action datasets.}\label{tab:ML_learning}
\end{table}
\section{Experiments}
Extensive experimental results on four public multi-view human action datasets are provided in this section. We compare the proposed CAM with several state-of-the-art methods, and provide a detailed ablation study to validate the effectiveness of each model component. Qualitative analyses of the attention distribution captured by CAM are also presented to further demonstrate the motivation of this work.

\subsection{Multi-View Action Datasets}


\emph{EV-Action}~\cite{wang2019ev} is a novel large-scale multi-view action dataset containing RGB, depth, skeleton, and electromyography (EMG) views. It contains 20 common human actions. 
We use the first 53 subjects for evaluation. Each subject performs each action 5 times so that we have 5300 samples in total. RGB and depth are used in our multi-view action recognition experiments. We choose the first 40 subjects as training set 
and the rest 13 subjects as test set.

\emph{NTU RGB+D}~\cite{shahroudy2016ntu} is a popular large-scale multi-view action dataset. It contains 56000 action clips in 60 action classes performed by 40 subjects. We choose the RGB and depth views for our experiments. We use the cross-subject evaluation strategy in the original dataset paper, which contains 40320 samples for training and 16560 samples for test.

\emph{UWA3D Multiview Activity \uppercase\expandafter{\romannumeral2} (UWA3D)} \cite{rahmani2016histogram, rahmani2014hopc} contains 30 human actions performed by 10 subjects. There are RGB, depth, and skeleton views recorded from 4 different viewpoints. We use the RGB and depth recorded from front view for evaluation. There are totally 270 samples available and we randomly choose 150 for training and 120 for test.

\emph{Depth-included Human Action Dataset (DHA)} \cite{lin2012human} is a multi-view dataset containing RGB and depth views. It contains 23 actions performed by 21 subjects. There are 483 video samples in total. We randomly choose 240 for training and the rest 243 for test. 

We test our model on both large (EV-Action and NTU RGB+D) and relatively small datasets (UWA3D and DHA) for a comprehensive evaluation.

\begin{figure*}[t]
\centering
\begin{center}
\includegraphics[width=1.0\linewidth]{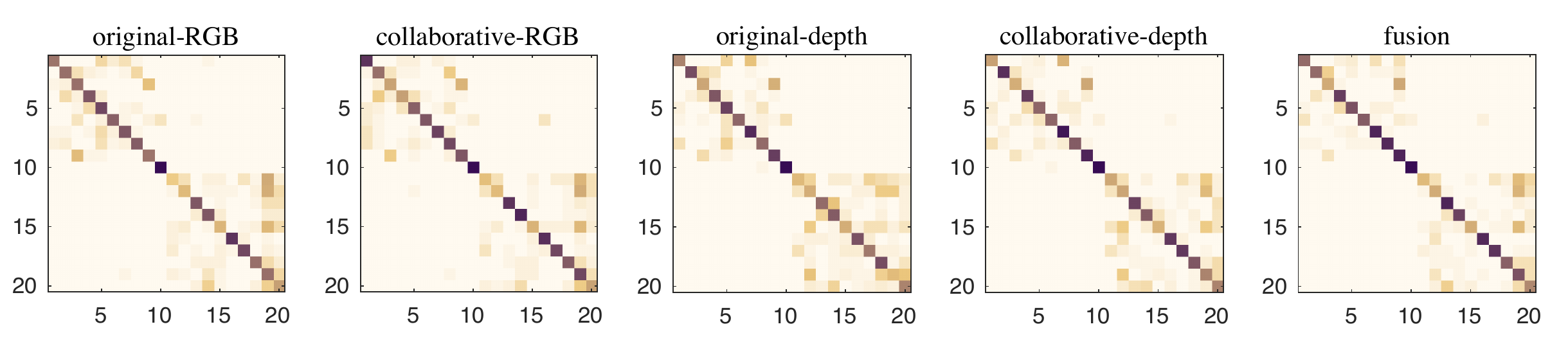}
\end{center}
\vspace{-5mm}
\caption{Visualization results of confusion matrices on EV-Action dataset.}
\label{fig:cm}
\vspace{-2mm}
\end{figure*}

\subsection{Baseline Methods}
For EV-Action and NTU datasets, we use several recent methods for comparison. 
For UWA3D and DHA datasets, we use several dataset specific baselines. We include the most recent state-of-the-art model for all datasets.

\begin{itemize}
\setlength\itemsep{0em}
\item MLSTM-FCN \cite{karim2019multivariate} is a novel deep framework proposed for handling multivariate temporal data. It contains a two-pathway structure (CNN and LSTM) to encode temporal data. Fully exploited patterns are captured for recognition.

\item RC framework \cite{bianchi2018reservoir} proposes a reservoir computing (RC) approach to model temporal data as vectorial representations in an unsupervised fashion. 
    
\item MFN \cite{zadeh2018memory} designs a memory fusion mechanism for multi-view learning based on temporal data. It proposes an early fusion strategy to integrate multi-view information in the feature space and improve the multi-view performance.

\item TSN \cite{wang2016temporal} is an effective benchmark model for action recognition. It utilizes an efficient sampling method and a two-stream structure to effectively collect valuable patterns and achieve promising performance. 
    
\item AMGL \cite{nie2016parameter} is a novel multi-view classification method based on graph learning. It aims to optimize weights for each graph automatically in a parameter-free fashion. 
    
\item MLAN \cite{nie2017multi} proposes an adaptive graph-based algorithm. It achieves the local structure and semi-supervised learning at the same time for multi-view learning. 
    
\item GMVAR \cite{wang2019generative} utilizes the generative strategy to mutually augment the multi-view representations. It boosts the multi-view learning performance significantly and improves the model robustness simultaneously. 
\end{itemize}

We use the first three methods for EV-Action amd NTU datasets, and the next four for UWA3D and DHA datasets. The GMVAR, as a state-of-the-art method for multi-view action recognition, is used for evaluation on all datasets.

\subsection{Implementation}
We use the same strategy to preprocess the raw videos for four datasets. Specifically, we use TSN \cite{wang2016temporal} to extract frame-level features for RGB view using the BNInception network as backbone. Each RGB frame is extracted into 1024 dimension feature vector. The depth view is transferred into RGB format first using HHA encoding algorithm \cite{gupta2014learning}. Then, we use the exactly the same TSN framework to extract depth features. We arrange the length of video with a unified number for each dataset via the cutting and repeating strategies. Concretely, for longer video, we pick the first certain frames and cut the rest video off; while for shorter video, we repeat the whole video sequence several times until it reaches the target number. The lengths of the sequence we set are 60, 60, 40, and 60 for EC-Action, NTU, DHA, and UWA3D, respectively. 

We concatenate the multi-view data in feature dimension to implement the MLSTM-FCN and RC classifier models. The MFN and GMVAR are designed for multi-view learning which fit our input data appropriately. TSN can be conducted for RGB and depth (transferred into RGB) individually without multi-view learning scenario. We adopt the AMGL and MLAN to fit our multi-view learning scenario. 

As shown on Fig.~\ref{fig:framework}, the view-specific attention is first trained individually. The input is multi-view action data. The attention distributions $Z^v$ are derived through optimizing Eq.~\ref{eq:view_specific_loss} during first-stage model. Next, the same action is set as input for the multi-view collaboration (second-stage). The MAR model is conducted with the additional input $Z^v$. Single view results from MAR model are fed into the final fusion model to obtain the multi-view performance. We set 128 batch size for EV-Action and NTU, and 32 for the other two datasets. The hidden dimensions for both temporal encoders (first and second stages) and attention are 128. The learning rates are 0.0005 and 0.001 for first-stage and second-stage. Our model is implemented by PyTorch with GPU acceleration.

\subsection{Performance Analysis}
The recognition performances of EV-Action and NTU are shown in Table~\ref{tab:ML_learning}. Our method outperforms all other approaches on both single-view and multi-view scenarios. MLSTM-FCN is an effective model for single view temporal data. It achieves competitive results. However, the fusion result is lower than ours. More importantly, our single-view performances are also higher which demonstrates our MAR model works well for discovering more valuable information. GMVAR is the state-of-the-art multi-view action recognition method based on generative model. However, it suffers from the difficulties of training generative model and cannot obtain promising performance on these two large-scale datasets.
In Table~\ref{tab:ML_learning}, we report the performances on DHA and UWA3D datasets. GMVAR achieves competitive performances on these two small-scale datasets. Its generative strategy improves the multi-view learning performance and model robustness. However, our method still generally outperforms it especially on single-view scenario. We visualize the confusion matrices on EV-Action test set containing the single-view results before/after our collaborative learning and the multi-view fusion result in Fig.~\ref{fig:cm}.


\begin{table}[t]
\begin{center}
\scalebox{1.0}{
\begin{tabular}{cccc}
\toprule
Method & RGB & Depth & RGB-D \\
\midrule
LSTM (baseline)     & 0.6878 & 0.6772 & - \\
CAM w/o MAR         & 0.6894 & 0.6796 & 0.7154 \\
CAM w/o RGB         & 0.6978 & 0.6802 & 0.7285 \\
CAM w/o Depth       & 0.6874 & 0.7084 & 0.7255 \\
CAM (ours)          & \textbf{0.7022} & \textbf{0.7123} & \textbf{0.7359} \\
\bottomrule
\end{tabular}}
\end{center}
\caption{Ablation Study on the EV-Action dataset.}\label{tab:ablation}
\end{table}

\subsection{Ablation Study}
We provide a detailed ablation study based on the EV-Action dataset to prove the necessity of each model component. The results are shown in Table~\ref{tab:ablation}. Particularly, we compare with with four ablated models as follows: 
1) \texttt{LSTM (baseline)} indicates single-view performance without multi-view collaboration learning and late fusion. 
2) \texttt{CAM w/o MAR} means we train the multi-view data synchronously and add late fusion without collaborative learning with the MAR cell. 
3) \texttt{CAM w/o Depth} and 4) \texttt{CAM w/o RGB} denote we only deploy collaboration learning to update the cell state of RGB and Depth view individually. 
We conclude deploying collaborative learning on each single-view enhances the representations and improves the performance correspondingly. Our complete model deploys the multi-view collaboration to achieve high performance on each single view. Further the late fusion leverages the enhanced single-view representations and obtains the best multi-view performance.
\begin{figure}[ht]
\centering
\begin{center}
\includegraphics[width=0.75\linewidth]{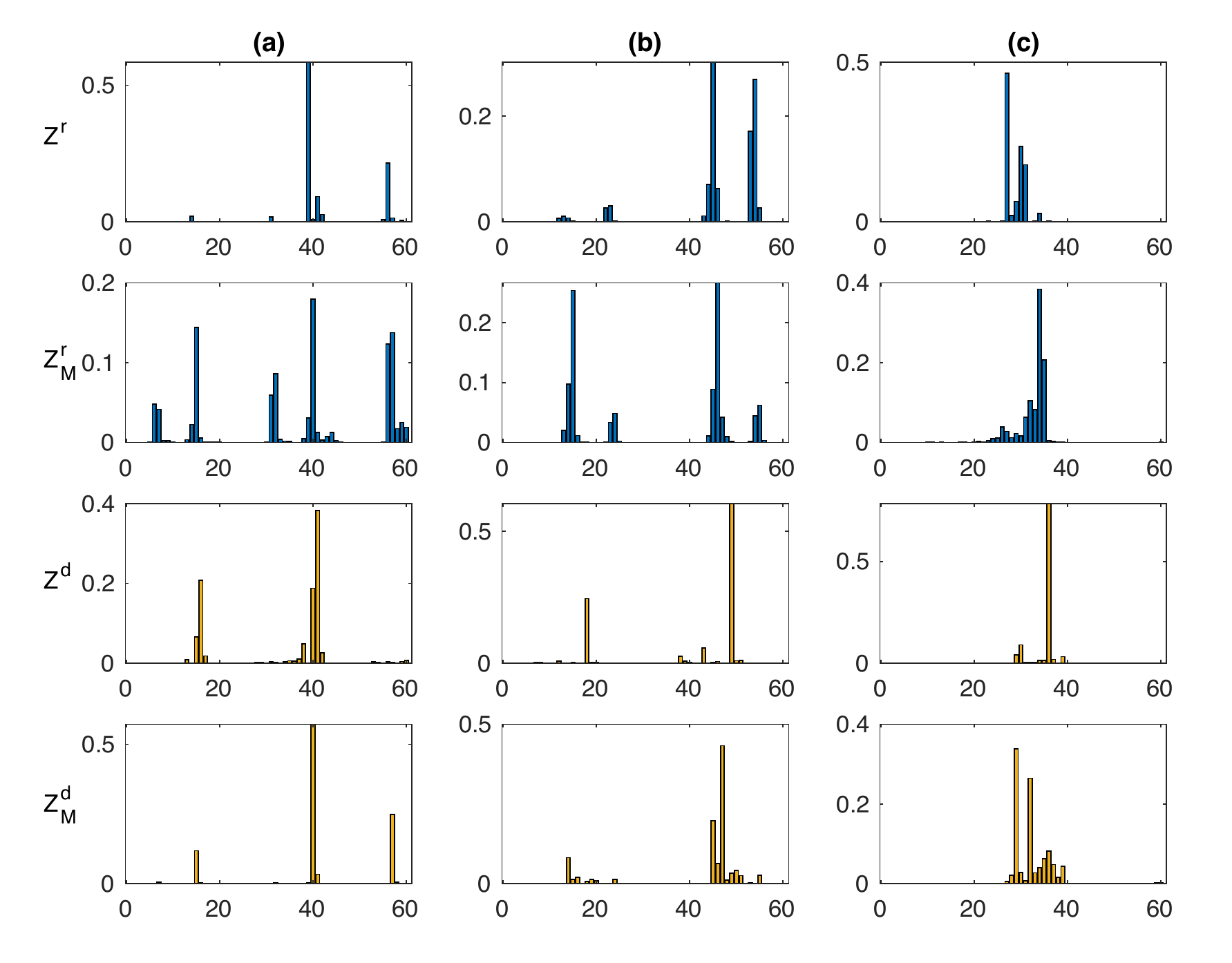}
\end{center}
\vspace{-2mm}
\caption{Visualization for the changes of attention distribution scores for two views. Each column represents one sample. X-axis is the time axis. Y-axis is the score values.}
\label{fig:three_sample}
\vspace{-2mm}
\end{figure}
\subsection{Attention Visualization}
We visualize and compare the changes between the $Z^v$ and $Z^v_M$, which are the temporal attention distributions (scores) before and after our multi-view collaboration model, respectively. It illustrates the collaborative learning process and provides the intuition about our model insight. Fig.~\ref{fig:three_sample} provides three samples from EV-Action dataset showing different collaborative learning cases. Each column represents one sample. In sample (a), each view captures specific attention patterns and guides the other view correspondingly. In sample (b), depth exerts an influence on RGB, while RGB has little impact on depth. In sample (c), two views roughly pay attention to the same location, however, they still adjust their attention scores through the collaborative learning process. 

\begin{figure*}[ht]
\centering
\begin{center}
\includegraphics[width=1.0\linewidth]{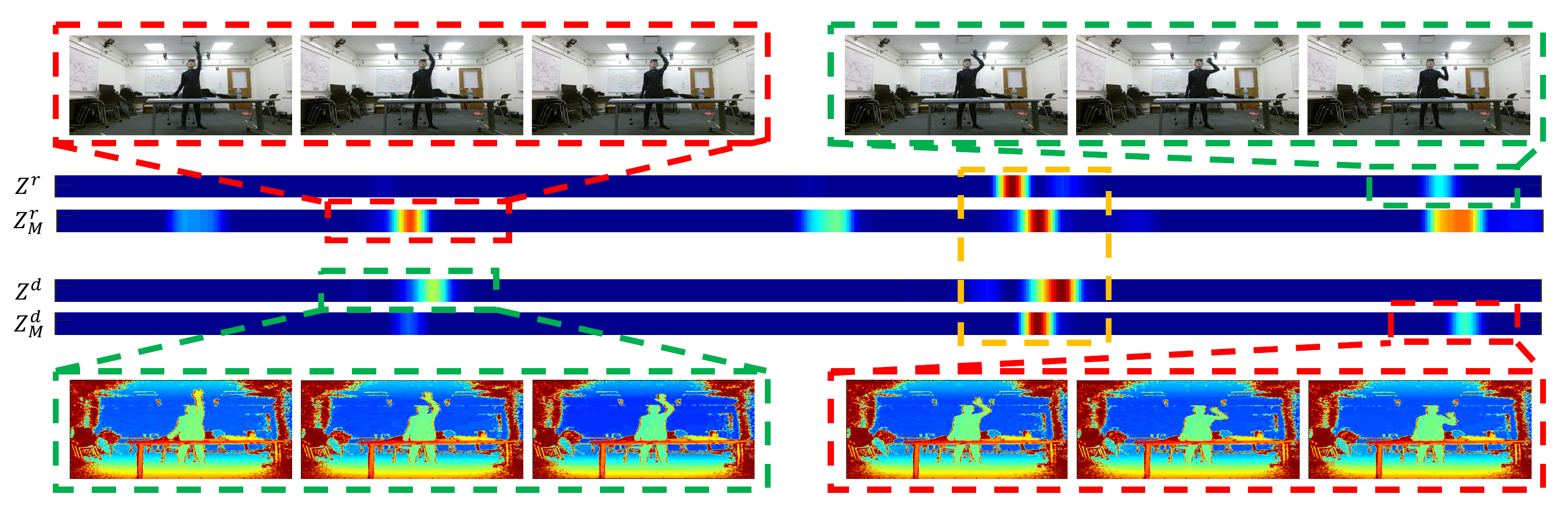}
\end{center}
\vspace{-3mm}
\caption{The colorbars represent the attention distribution scores of $Z^v$ and $Z^v_M$. The dash boxes indicate the temporal locations and contain its corresponding frames. Green indicates frames originally noticed by single view itself. Red represents frames attended after the collaborative learning. Yellow means frames noticed by both two views.}
\label{fig:one_sample}
\end{figure*}

Further, we provide more details of sample (a) with corresponding frames in Fig.~\ref{fig:one_sample}. The colorbars in the middle are the temporal attention scores of $Z^v$ and $Z^v_M$. Being Lighter means higher value. The green dash boxes indicate the frames have been noticed by each single view itself. The red boxes represent the frames gained attention after our collaborative learning, which is hard to be discovered by single view itself. The yellow boxes denote the frames noticed by both two views. In this case, the action class is ``throwing a ball''. The ``hands up'' and ``hands down'' are easily captured by RGB. While the ``throwing'' when hands at the highest point is easily noticed by depth view due to its motion changes in depth direction. The collaborative learning process takes advantages of characteristics of each view to guide the other view obtaining more valuable patterns and enhancing the learned representations. 
\section{Conclusions}
In this paper, we have proposed a Collaborative Attention Mechanism (CAM) for the multi-view action recognition task. A view-specific attention is first utilized for capturing multi-view attention distributions. Then, the multi-view collaboration is achieved with the proposed Mutual-Aid RNN (MAR) cell. In this way, each view is guided by the knowledge from the other view and enhanced to discover more latent information. The proposed CAM provides a novel perspective to leverage the attention mechanism and further explore multi-view sequential learning. The interpretability of attention is appropriately exploited to guide the learning process. Taking advantage of the collaboration strategy, the proposed CAM model outperforms state-of-the-art methods on four publich action datasets in both single and multi-view scenarios. A detailed ablation study has been provided to validate the effectiveness of each model component.

\bibliographystyle{plainnat}
\bibliography{arxiv}
\end{document}